\documentclass[10pt,twocolumn]{article}
\usepackage[T1]{fontenc}
\usepackage[utf8]{inputenc}
\usepackage[letterpaper,top=0.68in,bottom=0.74in,left=0.66in,right=0.66in,columnsep=0.22in]{geometry}
\usepackage{times}
\usepackage{microtype}
\usepackage{amsmath,amssymb}
\usepackage{graphicx}
\usepackage{booktabs,tabularx,array,adjustbox}
\usepackage{longtable}
\usepackage{enumitem}
\usepackage[font=small,labelfont=bf]{caption}
\usepackage[numbers,sort&compress]{natbib}
\usepackage{xcolor}
\usepackage[hidelinks]{hyperref}
\usepackage{xurl}
\setlist{nosep,leftmargin=*}
\providecommand{\tightlist}{\setlength{\itemsep}{0pt}\setlength{\parskip}{0pt}}

\title{Verification-Guided Retrosynthesis: Reaction Feedback and Evidence Timing}
\author{Hongxuan Shi \quad Hankz Hankui Zhuo\thanks{Corresponding author.}\\[0.4em]
School of Artificial Intelligence, Nanjing University, China\\[0.2em]
\texttt{502025370040@smail.nju.edu.cn} \quad \texttt{hankz@nju.edu.cn}}

\date{}
\begin{document}
\maketitle
\begin{abstract}
Retrosynthesis planning seeks synthesis routes from purchasable starting materials to a target molecule. Search over predicted reactions can connect a target to available materials, yet an infeasible reaction can invalidate the resulting route. Evaluating route quality only after search has finished cannot correct errors made during search. We introduce Verification-Guided Reflow (VGR), incorporating reaction verification as a decision action in AND/OR search. Verification results are represented as binary verdicts; queries are scheduled from the search structure, and acquired verdicts exclude reactions or adjust search priorities. We define verified solvability as the fraction of targets for which a complete route accepted stepwise by the verifier is found. On Retro190, feeding verification results into search increases verified solvability from 51.1\% to 62.6\% relative to route evaluation without feedback. Further controlled experiments show that earlier verification and immediate prioritization of accepted reactions each improve route discovery, yet combining them performs worse than either change alone. We therefore propose early pruning and delayed guidance: rejected reactions are pruned immediately, while accepted reactions gain search priority when processed as part of a complete candidate route. In a separate comparison with the verification scheduling rule fixed, this strategy increases verified solvability from 61.1\% with immediate guidance to 71.1\%, using fewer proposal calls and unique verifier queries.
\end{abstract}
\section{Introduction}\label{introduction}

Multistep retrosynthesis planning starts from a target molecule and searches backward for synthesis routes from purchasable starting materials. Neural one-step models and search algorithms can already find complete routes for many targets within a computational budget \citep{segler2018planning, chen2020retrostar}. Such success first means that every route leaf reaches purchasable materials, which we call topological solvability. Reactions in the graph are synthesis choices proposed by a model: even a topologically complete route can fail as a synthesis plan if one reaction is infeasible. A high topological solve rate therefore does not imply that the planner has found equally many feasible synthesis plans.

We use forward reaction prediction to check whether each step produces its intended product. A target is verified-solvable if search finds at least one complete route whose reactions all pass verification; verified solvability is the fraction of targets satisfying this condition. On Retro190, the planner reports 90.5\% topological solvability, but verified solvability is only 30.5\% when the first complete route is returned by default. Thus, about two thirds of targets reported as solved have a default route that fails stepwise verification. Even if search continues to the budget limit and repeatedly checks complete candidates, verified solvability reaches only 51.1\% when verdicts do not change search. The question is therefore not simply how to find more routes, but how to let discovered feasibility problems change subsequent exploration.

This gap has a search-level explanation. For a molecule \(x\) to be synthesized, a one-step model proposes reaction \(R\) with probability \(p(R\mid x)\). In retro*, search converts this probability into reaction cost \(-\log p(R\mid x)\) and prioritizes choices favored by the model. This score determines the order of exploration, while reaction feasibility requires checking the specific transformation. If search continues under its original ranking, a check can reject a discovered plan but cannot use that verdict to reallocate exploration. For verification to help route discovery, its verdicts must change subsequent search decisions.

Existing work supplies feasibility information through reaction filters, forward prediction, and route evaluation; Active Retrosynthetic Planning (ARP) further learns a policy for querying reaction quality \citep{kreutter2023ttl, liu2024synthesizability, yuan2024arp}. Verifying every candidate makes cost grow with the candidate pool, whereas evaluating only assembled routes remains constrained by the plans that the original search discovers. We study the following question: under limited computation, which reactions should be verified from the current search graph, and how should their verdicts be used to find more routes that pass stepwise verification?

We formulate this problem as AND/OR search with both expansion and verification actions. Molecule nodes represent alternative synthesis choices (OR), while reaction nodes require every precursor to be synthesized (AND). Expansion calls the one-step model to generate candidate reactions and their precursors for a molecule; verification acquires a binary verdict \(v(R)\in\{0,1\}\), where zero and one denote rejection and acceptance. A verification schedule selects queries using reactions' positions in the graph and the completion state of routes. A feedback rule determines how verdicts constrain reaction choices and change search costs. Through subsequent expansions, these decisions jointly affect route discovery. Our objective is to improve verified solvability under expansion and search-query budgets, delivering only complete routes with an acquired positive verdict for every reaction.

Verification-Guided Reflow (VGR) implements this formulation. Reflow means persistently updating search state with verification verdicts so that they affect subsequent reaction choices and route priorities. A negative, \(v(R)=0\), excludes the reaction; a positive, \(v(R)=1\), can lower its search cost and guide the planner toward completing a route containing it. Three policies all check reactions after a complete candidate route forms and differ in how they use verdicts: post-hoc checks candidates without feedback, prune excludes negatively verified reactions, and guide additionally uses positives to guide search. With the same proposal model, verifier, and budgets, their verified solvability on Retro190 is 51.1\%, 59.5\%, and 62.6\%, respectively, while their topologically solved target sets are identical. The two forms of feedback thus help the planner find more stepwise-verified routes among targets it can already solve topologically.

Can earlier verification and immediate, stronger guidance improve search further? Starting from verification at complete-candidate formation with negative-only pruning, we separately move a reaction's check to the first expansion of one of its non-stock precursors, or strengthen guidance by reducing accepted reactions' costs to zero. Under common budgets, either change improves coverage, yet their combination performs worse than either change alone.

The key to this interaction is the different meaning of the two verdict signs for a route. One negative suffices to exclude a candidate containing that reaction, while one positive supports only a local transformation; the remaining steps still need completion or verification. Strengthening a locally accepted reaction too early can concentrate search on a branch whose precursor synthesis remains unresolved. Guidance also changes subsequent expansions and hence the reactions available for later verification. When a verdict is acquired and when it takes effect therefore require separate control and joint investigation.

We consequently propose early pruning and delayed guidance. Negatives immediately exclude reactions, while positives are retained. Once search forms a complete candidate from the target to purchasable materials, a positive changes its reaction's priority when that reaction is processed during the candidate's stepwise evaluation. Failure information can prevent wasted investment early, while positive information guides search once the structure of a complete route is available. In a separate comparison without a search-query ceiling, holding the verification schedule and guidance strength fixed, this strategy increases verified solvability from 61.1\% with immediate guidance to 71.1\%, using fewer one-step calls and verifier queries. Monte Carlo tree search controls also show a benefit from delayed guidance, while a different one-step proposal model preserves the post-hoc--prune--guide ordering. Expert labels and cross-model route evaluation provide external evidence for the chemical relevance of the verification criterion.

Our two contributions are:

\begin{enumerate}
\def\labelenumi{\arabic{enumi}.}
\tightlist
\item
  \textbf{Reaction verification as an action in retrosynthesis search, with controlled evaluation of its feedback.} We jointly describe verification scheduling and search-state updates in an AND/OR graph. The post-hoc--prune--guide comparison identifies the contributions of negative exclusion and positive guidance to verified solvability.
\item
  \textbf{The interaction between verification scheduling and feedback timing, with early pruning and delayed guidance.} Motivated by the failure of directly combining early verification with strong guidance, we distinguish verdict acquisition from guidance activation and use fixed-schedule comparisons to test the benefit of sign-asymmetric timing.
\end{enumerate}

\section{Route Discovery with Reaction Verification}\label{route-discovery-with-reaction-verification}

\subsection{Candidate routes and observable evidence}\label{candidate-routes-and-observable-evidence}

Fix a target molecule \(g\), purchasable inventory \(\mathcal S\), one-step proposal model with parameters \(\theta\), and native search algorithm. Expanding molecule \(x\) returns a finite set of reactions \(\mathcal A_\theta(x)\). Molecules are OR nodes because one synthesis choice suffices; reactions are AND nodes because every precursor must reach stock. At planner-update index \(t\), the exposed finite graph is \(G_t\). The index counts state updates; one-step-model calls are accounted for separately. A target-to-stock solution subgraph is a \textbf{proof} during search, and its extracted reaction tree is a \textbf{route}. It is \textbf{topologically closed} when all leaves belong to \(\mathcal S\).

A fixed verifier with parameters \(\phi\) and measurement protocol \(\mathcal M\) assigns \(v_{\phi,\mathcal M}(R)\in\{0,1\}\) to reaction \(R\). For a fixed target, identical reaction keys have consistent verdicts; the key definition appears in Appendix A. The evidence record \(Z_t\) contains acquired verdicts, while unqueried reactions remain unknown. For a nonempty, topologically closed route \(\rho\), define its certificate by

\[
C_{\mathcal M}(\rho)=\prod_{R\in\rho}v_{\phi,\mathcal M}(R).
\]

The planner delivers \(\rho\) only after all its verdicts have been acquired and \(C_{\mathcal M}(\rho)=1\). Thus a reaction observation and a complete route certificate are different objects. Expert-label and cross-model evaluations assess the external chemical signal in this operational endpoint.

\subsection{Acquisition and feedback as coupled decisions}\label{acquisition-and-feedback-as-coupled-decisions}

A policy \(\pi=(q,w)\) augments the fixed native searcher. Acquisition rule \(q\) reads \((G_t,Z_t)\) and the remaining budget to select reactions for verification. Writeback rule \(w\) uses the current graph and acquired evidence to determine action masking or priority changes, including when these updates take effect. The native searcher then chooses its next expansion from the updated state. Acquisition therefore changes the evidence available to feedback, and feedback changes the graph on which acquisition acts.

Let \(\mathcal D\) be the target set and \(S_g^{q,w}(N,M_{\mathrm{search}})\in\{0,1\}\) indicate whether the policy delivers a certified route for target \(g\) within the specified budgets. The policy-design objective is high certified coverage,

\[
J(q,w;N,M_{\mathrm{search}})=\frac{1}{|\mathcal D|}
\sum_{g\in\mathcal D}S_g^{q,w}(N,M_{\mathrm{search}}),
\]

subject to at most \(N\) one-step calls and, where imposed, \(M_{\mathrm{search}}\) unique search-time reaction queries per target. The uncapped-search-query setting is denoted \(M_{\mathrm{search}}=+\infty\). Delivery checks are accounted for outside this search-query ceiling. Total physical verification cost is the number of distinct target-local keys queried across search and delivery; cached reuse and positive-credit release consume no new query. Appendix D gives the separate logical-event accounting used in the acquisition-frontier experiment.

The experiments manipulate \(q\) and \(w\) while fixing the proposal model, verifier, and search backend. Post-hoc keeps verdicts out of planner state, so its graph trajectory \(G_t^0\) is verdict-blind even when several complete candidates are checked. Prune changes action availability; guide additionally changes priority. Their coverage comparison measures the benefit of feeding the same criterion back into discovery. We then cross two acquisition rules with two feedback strengths to measure whether these interventions compose additively, and hold acquisition fixed to isolate positive-credit timing.

\subsection{Why the two signs have different implications}\label{why-the-two-signs-have-different-implications}

The AND/OR structure gives a simple distinction. If \(v_{\phi,\mathcal M}(R)=0\), every route containing \(R\) has certificate zero. If \(v_{\phi,\mathcal M}(R)=1\), the route still needs all other reaction verdicts to be positive and all precursor obligations to reach stock. Negative evidence can therefore exclude a candidate immediately under the fixed criterion, while useful positive guidance also depends on the surrounding search state. Section 3 turns this distinction into separately controllable updates and states the properties they preserve.

\section{Verification-Guided Reflow}\label{verification-guided-reflow}

VGR separates a verdict's acquisition, its effect on search, and the final certification of a route. Evidence is retained in a target-local cache; the native planner maintains the graph and selects expansions after each writeback. We first define the update interface, then the structural query and release rules.

\subsection{Persistent feedback and positive-credit strength}\label{persistent-feedback-and-positive-credit-strength}

In additive-cost search, reaction \(R\) proposed for molecule \(x\) has native cost \(c_0(R)=-\log p_\theta(R\mid x)\), with retained proposals satisfying \(0<p_\theta(R\mid x)\le1\). After an enabled negative update, its effective cost becomes \(+\infty\) and the native AND/OR propagation invalidates dependent candidates. For an acquired positive, let \(r_t(R)\in[0,1]\) denote the released fraction of its positive search credit. Its cost is

\[
c_t(R)=\bigl[1-(1-\alpha_{\min})r_t(R)\bigr]c_0(R),
\qquad 0\le\alpha_{\min}\le1.
\]

Unknown reactions retain \(c_0\). For a positive, \(r_t=0\) also retains \(c_0\), and \(r_t=1\) sets the cost to \(\alpha_{\min}c_0\). Thus \(\alpha_{\min}=1\) is positive-neutral and \(\alpha_{\min}=0\) is maximal credit. Strength and release time are separate controls. Post-hoc writes neither sign into navigation; prune applies negative updates only; guide also releases positive credit and propagates the new priorities. All verdicts used for feedback remain available for later route checks.

For prior-guided Monte Carlo tree search (PUCT), negative verdicts mask actions, while a released positive changes its prior rather than its backed-up value. If \(P_0(s,a)\) is the native prior of action \(a\) at search state \(s\), the unnormalized released weight is \(P_0(s,a)^{\alpha_{\min}}\) for \(\alpha_{\min}>0\) and one at \(\alpha_{\min}=0\), followed by sibling renormalization. Native visit counts and value backup remain unchanged. Appendix B supplies the update and endpoint details.

\subsection{Structural acquisition and sign-asymmetric release}\label{structural-acquisition-and-sign-asymmetric-release}

\textbf{Closure-time acquisition} checks reactions while processing the current minimum-effective-cost, topologically closed root candidate. This concentrates measurements on reactions already assembled into a complete target-to-stock plan. \textbf{Post-investment acquisition (PI)} moves a parent reaction's check to the first one-step call spent on one of its non-stock precursor children. PI retains closure-time completion checks for unknown reactions on a complete candidate. Both rules choose queries using the reaction's role in the evolving search graph.

Holding PI acquisition and immediate negative pruning fixed, we compare three positive-release rules:

\begin{table}[t]
\centering
\caption{Positive-evidence release controls. Acquisition and negative activation are separate from the release rule.}
\label{tab:release-controls}
\small
\setlength{\tabcolsep}{3pt}
\begin{tabularx}{\columnwidth}{@{}l c X@{}}
\toprule
\textbf{Positive release} & \textbf{$r_t(R)$} & \textbf{Effect on search} \\
\midrule
neutral & $0$ & retain native priority \\
immediate & $1$ & release credit upon acquisition \\
root-delayed & $h_t^{\mathrm{root}}(R)$ & release while processing a closed root candidate \\
\bottomrule
\end{tabularx}

\end{table}

The one-way gate \(h_t^{\mathrm{root}}(R)\) starts at zero and becomes one when the positively verified reaction occurrence is processed on an extracted, topologically closed proof from the target to stock. The candidate is processed in graph-creation order; a negative stops that attempt and triggers reselection, while already released positives remain active. Structural completion triggers release before all other reactions have necessarily passed verification. A failed candidate can therefore contribute reusable positive credit from its accepted prefix. This gate concerns a complete root candidate, not closure of a local reaction subtree.

VGR then checks the delivery candidate using cached verdicts and any remaining queries. It returns the route only if every step is positive; otherwise search continues within the proposal budget. A release reads an existing verdict and changes priority without another verifier call. Appendix B gives pseudocode and the separate read-only delivery procedure.

\subsection{Preserved properties}\label{preserved-properties}

\textbf{Criterion-relative pruning.} Fix an exposed graph \(G\) and consistent verifier verdicts, and let \(\mathcal F_{\mathcal M}(G)\) be its nonempty closed routes with certificate one. If reaction occurrence \(R\) has verdict zero, deleting that action preserves this set:

\[
\mathcal F_{\mathcal M}(G\setminus R)=\mathcal F_{\mathcal M}(G).
\]

Every route removed by the deletion contains \(R\) and already has certificate zero; every other route remains. This is a property of pruning in the currently exposed graph. Future graph exposure and finite-budget coverage depend on subsequent search and are measured experimentally.

\textbf{Certified delivery and query reuse.} The delivery condition ensures that every returned route is closed and every included reaction has an acquired positive verdict under \(\mathcal M\). Cached evidence can be reused across candidate attempts, and releasing a cached positive adds no measurement. Acquisition enforces the stated search-query ceiling; delivery cost is recorded separately.

These properties explain what the controller preserves while changing navigation. Whether positive feedback helps, and when it should act, are the empirical questions addressed by the control ladder, factorial experiment, and fixed-acquisition release comparison. Continuous local-maturity rules remain an additional exploration of this interface, with their definitions and complete results in Appendices B and E.

\section{Evaluation Protocol}\label{evaluation-protocol}

\subsection{Primary setup}\label{primary-setup}

Primary experiments use all 190 Retro190 targets \citep{chen2020retrostar}, full stock, top-50 reactions from a template-based multilayer perceptron (MLP), retro*, and at most \(N=500\) one-step calls per target. A fixed ReactionT5 model \citep{sagawa2023reactiont5} generates five products from reactants; a positive requires the target product after canonicalization without stereochemical markers. Search receives only this binary verdict. Target-local reaction keys reuse cached inference.

The post-hoc--prune--guide ladder and the closure/PI by neutral/maximal-credit factorial experiment use a search-query ceiling of \(M_{\mathrm{search}}=100\). The latter holds the ceiling fixed while changing acquisition and positive feedback. The release-clock comparison instead fixes PI acquisition, immediate negative pruning, and \(\alpha_{\min}=.25\), with no active search-query ceiling. It compares neutral, immediate, and root-delayed positive release. These experiments isolate different decisions; their endpoints are reported with their own budgets.

\subsection{Transfer, resources, and external evidence}\label{transfer-resources-and-external-evidence}

RetroChimera proposals and Syntheseus PUCT test transfer across proposal and search interfaces \citep{maziarz2024retrochimera, maziarz2025syntheseus}. PUCT compares neutral, immediate, and root-delayed priors; the main text uses maximal credit as a mechanism stress test, while Appendix E reports both \(\alpha_{\min}=.25\) and \(0\). A second task uses 100 target-only molecules derived from the DESP Pistachio Hard pairs; the source starting-material constraint is not retained \citep{yu2024desp}. The release controls and continuous local-maturity explorations on this task are reported in Appendix E.

Shared unique-query caps of 100 and 500, plus closure, PI, and full-on-use acquisition, characterize proposal--verification trade-offs. Full-on-use queries a reaction immediately before the first downstream call. URSA and FREA provide expert-labeled reaction audits \citep{zagribelnyy2026ursa, yu2026frea}. A fixed-route cross-model audit uses a forward model absent from online route selection. Appendix C gives policy definitions and configurations; Appendices D and E give accounting, statistics, and supporting results.

\subsection{Metrics and statistics}\label{metrics-and-statistics}

Verifier-certified success requires a closed route whose every step is accepted; topological success ignores verdicts. We separately report one-step calls, logical feedback applications, target-local unique queries, delivery checks, node visits, and PUCT simulations. Paired success uses exact McNemar tests, while effect uncertainty uses paired target bootstrap and product-cluster bootstrap for FREA. Appendix D gives the estimands, multiplicity handling, exact tests, and random seeds.

\section{Results}\label{results}

\subsection{From route checking to negative and positive feedback}\label{from-route-checking-to-negative-and-positive-feedback}

\begin{table*}[t]
\centering
\caption{Post-hoc, prune, and guide on Retro190 under $N=500$ and $M_{\mathrm{search}}=100$. The first-proof row records the initial topological route before continued post-hoc checking.}
\label{tab:interface-ladder}
\small
\begin{adjustbox}{max width=\textwidth}
\begin{tabular}{lrrr}
\toprule
\textbf{Policy} & \textbf{Topological} & \textbf{Verifier-certified} & \textbf{Mean one-step calls} \\
\midrule
topology-only first proof & 172 & 58 & 110 \\
post-hoc & 172 & 97 & 302 \\
prune & 172 & 113 & 277 \\
guide & 172 & 119 & 261 \\
\bottomrule
\end{tabular}
\end{adjustbox}
\end{table*}

Continuing the verdict-blind trajectory raises first-route certification from 58 to 97. Feeding negatives into search raises it to 113; adding positive credit raises it to 119. All three policies topologically solve the same 172 targets. Under the common \(N=500\) and \(M_{\mathrm{search}}=100\) limits, the post-hoc--prune--guide ladder shows how feedback changes supported-route discovery, with mean one-step calls decreasing from 302 to 277 and 261.

\subsection{Earlier querying and stronger guidance do not compose additively}\label{earlier-querying-and-stronger-guidance-do-not-compose-additively}

\begin{table*}[t]
\centering
\caption{Acquisition timing and positive-credit strength interact on all 190 Retro190 targets. Cells report verifier-certified / topologically solved targets under $N=500$ and $M_{\mathrm{search}}=100$.}
\label{tab:timing-credit}
\small
\begin{adjustbox}{max width=\textwidth}
\begin{tabular}{lrr}
\toprule
\textbf{Acquisition} & \textbf{Neutral positives} & \textbf{Maximal positive credit} \\
\midrule
root-proof closure & 113 / 172 & 119 / 172 \\
post-investment & 116 / 168 & 109 / 153 \\
\bottomrule
\end{tabular}
\end{adjustbox}
\end{table*}

Cells report certified/topological targets under \(N=500\) and \(M_{\mathrm{search}}=100\). Earlier acquisition improves negative-only coverage from 113 to 116; maximal credit improves closure-time coverage from 113 to 119. Their combination falls to 109/153.

To measure this interaction, let \(q_C\) and \(q_E\) be closure and PI acquisition, and \(w_P\) and \(w_G\) negative-only and maximal-credit writeback. Using the coverage \(J\) defined in Section 2, with common budgets suppressed, define

\[
I=\bigl[J(q_C,w_G)-J(q_C,w_P)\bigr]
-\bigl[J(q_E,w_G)-J(q_E,w_P)\bigr].
\]

Additive effects give \(I=0\); positive \(I\) means guidance helps more under closure-time acquisition. The certified interaction is \(13/190\), with uncertainty spanning zero. Replacing certified success with topological success gives an interaction of \(15/190\), or \(+7.9\) points (95\% confidence interval \([+4.2,+12.1]\)). The factorial comparison thus motivates a direct test of the positive-release clock while acquisition is held fixed.

\subsection{Early negative pruning with root-delayed positive credit}\label{early-negative-pruning-with-root-delayed-positive-credit}

\begin{table*}[t]
\centering
\caption{Positive-credit release under fixed PI acquisition, $N=500$, and $\alpha_{\min}=.25$, without an active search-query ceiling.}
\label{tab:release}
\small
\begin{adjustbox}{max width=\textwidth}
\begin{tabular}{lrrrr}
\toprule
\textbf{Positive release} & \textbf{Topological} & \textbf{Certified} & \textbf{$\Sigma$ one-step calls} & \textbf{Unique search queries} \\
\midrule
neutral & 165 & 126 & 44,471 & 46,089 \\
immediate & 152 & 116 & 48,758 & 49,428 \\
root-delayed & 165 & 135 & 42,579 & 43,974 \\
\bottomrule
\end{tabular}
\end{adjustbox}
\end{table*}

These controls share PI acquisition, immediate negative pruning, \(N=500\), and \(\alpha_{\min}=.25\), without an active search-query ceiling. Root-delayed versus immediate release rescues 25 and loses 6 targets: \(+10.0\) points (95\% CI \([4.7,15.8]\)), with 12.7\% fewer one-step calls and 11.0\% fewer unique search queries. Root-delayed and neutral have identical target-wise topology and identical one-step call counts at first closure, yet root-delayed release certifies 135 versus 126 targets (11 rescues and 2 losses). Positive credit therefore adds value after structural completion beyond the benefit of early negative pruning.

\subsection{Transfer and external chemical evidence}\label{transfer-and-external-chemical-evidence}

In PUCT, the maximal-credit stress test gives 137 certified targets under root-delayed release versus 125 under immediate release (13 rescues and one loss), with 35.0\% fewer simulations and 62.0\% fewer nodes. With milder credit (\(\alpha_{\min}=.25\)), the corresponding coverage is 136 versus 132. The complete strength matrix and paired success-curve area statistics are in Appendix E. RetroChimera proposals preserve the post-hoc--prune--guide direction (162/174/180), testing feedback through a stronger proposal interface.

On the 100-target Pistachio-derived task, post-hoc certifies 61 targets, neutral release 64, and immediate and root-delayed release 66 each. Appendix E reports the full comparison and the continuous local-maturity explorations on both target sets.

The certificate has external chemical signal. ReactionT5 reaches 77.1\% balanced accuracy on 1,000 expert-labeled URSA reactions and 71.4\% on evaluable FREA reactions. In a fixed-route audit selected from 251 targets successfully solved by guide with an online Graph2SMILES verifier \citep{tu2022graph2smiles}, a condition-aware ReactionT5 model absent from route selection accepts 181/251 guide routes versus 117/251 paired first-topological routes. This is a paired cross-model comparison within the guide-success cohort. Appendix E gives the cohort definition, coverage, and complete expert-label audit.

\section{Related Work}\label{related-work}

Modern retrosynthesis combines one-step models with best-first AND/OR search, learned values, or Monte Carlo tree search \citep{segler2018planning, chen2020retrostar, genheden2020aizynthfinder, hong2023egmcts, liu2023pdvn, maziarz2025syntheseus}. VGR places a verifier-mediated control layer between reaction proposal and search.

Feasibility models support eager filtering, round-trip checks, and route evaluation \citep{kreutter2023ttl, sadowski2025retrotrim, liu2024synthesizability}. ARP learns a reaction-conditioned acquisition policy \citep{yuan2024arp}. VGR's structural rules instead select checks using downstream search investment and complete-candidate participation, and separate acquisition from the activation of each verdict sign. Retro-fallback addresses portfolios under prior reaction uncertainty \citep{tripp2024retrofallback}, whereas VGR studies action and priority updates after observations arrive.

LazySP and Generalized Lazy Search evaluate expensive edges, update state, and replan \citep{dellin2016lazysp, mandalika2019gls}. VGR studies these decisions in an online-generated AND/OR synthesis graph, where a positive observation supports one reaction but leaves precursor obligations unresolved. Its controlled comparisons identify the acquisition--writeback interaction and test different positive-release clocks.

\section{Discussion}\label{discussion}

Reaction verification can change route discovery by changing which actions remain available and which branches receive further search. The control ladder isolates these roles. The factorial experiment then shows why adding earlier acquisition to positive guidance requires a separate design decision: an acquired positive concerns one reaction, whereas the usefulness of developing that branch depends on the remaining precursor obligations.

Root-delayed release uses an observable structural event to distinguish local support from a complete candidate. It preserves early negative pruning while letting accepted reactions influence later search once they have participated in a root candidate. Positives released before a later failure remain available to alternative proofs. Fixed-PI comparisons and PUCT controls measure the benefit of this release clock through different navigation interfaces.

Acquisition remains resource-dependent. At \(\alpha_{\min}=.25\), under a shared unique-query ceiling of 100, closure acquisition leads post-investment root-delayed release 126--113; at 500, the order reverses to 135--128. Full-on-use acquisition can reach 144 certifications but spends 2.26 times post-investment acquisition's unique queries and loses topology. We report proposal calls and verifier work separately; a matched total search-plus-delivery query budget and dedicated wall-clock measurements would further characterize resource efficiency (Appendices D and E).

Continuous local-maturity rules provide an additional way to vary positive-credit release. Appendices B and E give their definitions, complete parameter grid, and results.

External expert-label and cross-model audits support the chemical relevance of the certificate. Reaction conditions, stereochemistry, isolation, and wet-lab execution provide further evidence channels beyond the reported protocol. Search control improves the discovery of supported plans; connecting these plans to experimental route selection is a natural next stage.

\section{Conclusion}\label{conclusion}

A complete predicted synthesis route can still fail chemical checks, making the use of verification during discovery important. VGR retains reaction verdicts and propagates their effects into AND/OR search. The post-hoc--prune--guide ladder establishes the value of negative and positive feedback, while factorial and fixed-acquisition controls expose the interaction between evidence timing and guidance. Immediate negative pruning with root-delayed positive credit provides a concrete response to that interaction. These findings make the selection and activation of reaction evidence explicit design variables for verified route discovery.
\small
\bibliographystyle{plainnat}
\bibliography{references}
\clearpage
\appendix
\renewcommand{\thesection}{\Alph{section}}
\section{Operational Semantics}\label{operational-semantics}

The main paper treats verification as a state-changing search operation. This supplement fixes the implementation semantics needed to reproduce that operation without conflating reaction identity, verifier matching, graph occurrence, and delivery certification.

\subsection{Reaction identity, cache scope, and counters}\label{reaction-identity-cache-scope-and-counters}

SMILES is a molecular string representation. For each target, a reaction key is the planner product SMILES followed by \texttt{\textgreater{}\textgreater{}} and the lexicographically sorted planner reactant SMILES joined by dots. Exact planner strings, including stereochemical markers present in planner state, define cache identity. Stereochemistry is removed only by the fixed verifier's product-matching rule. One target-local key receives one verdict, and every occurrence of that key reuses the verdict. Caches are never shared across targets in the reported scientific comparisons.

The implementation separates four counts:

\begin{itemize}
\tightlist
\item
  \(n_{\mathrm{search}}\): logical verdict applications that can modify search;
\item
  \(n_{\mathrm{search}}^{\mathrm{uniq}}\): target-local unique keys physically queried for search feedback;
\item
  \(n_{\mathrm{certify}}\): logical route-step checks during delivery;
\item
  \(n_{\mathrm{total}}^{\mathrm{cache}}\): unique keys in the union of search and delivery checks.
\end{itemize}

The search ceiling \(M_{\mathrm{search}}\) applies only to \(n_{\mathrm{search}}^{\mathrm{uniq}}\). Cache hits do not consume new search-query quota, and delivery certification is accounted for outside this ceiling. The cap experiments therefore equalize the maximum opportunity for search-time feedback; they do not equalize realized verifier inference or impose a hard cap on total search-plus-delivery queries. Releasing a cached latent positive changes planner state without invoking the verifier again.

The acquisition-frontier runs reported below use a different cap, \(M_{\mathrm{event}}\), on logical search-feedback events. Cached reuse consumes event quota there. The shared-ceiling comparisons instead cap physically unique target-local search queries with \(M_{\mathrm{search}}\). These budget units define different interventions and are not treated as interchangeable.

\subsection{State-transition invariants}\label{state-transition-invariants}

The implementation enforces six invariants.

\begin{enumerate}
\def\labelenumi{\arabic{enumi}.}
\tightlist
\item
  \textbf{Target-local key consistency.} One key receives one reusable verdict.
\item
  \textbf{Immediate negative activation.} An acquired search-feedback negative masks the affected reaction action and propagates through dependent proofs. The additive tree implementation reuses the target-local verdict when processing another occurrence; PUCT additionally masks existing and future occurrences of the key.
\item
  \textbf{Navigation-neutral latent positives.} Before release, a positive retains the original proposal cost or PUCT prior.
\item
  \textbf{Zero-query release.} A transition from latent positive evidence to released positive evidence reads the cache and consumes neither another verifier call nor feedback quota.
\item
  \textbf{Monotone transitions.} A pruned reaction is not restored, and a released positive does not revert to latent or unknown.
\item
  \textbf{Stepwise delivery certification.} Every reaction in a returned route has a positive verdict under the fixed measurement protocol.
\end{enumerate}

\section{Reference Procedures}\label{reference-procedures}

\subsection{Three-clock controller}\label{three-clock-controller}

The following operational outline makes acquisition, negative activation, and positive release explicit, using PI's post-expansion order. FU checks its acquisition trigger before step 2 and may block that expansion; post-hoc bypasses search writeback.

\begin{enumerate}
\def\labelenumi{\arabic{enumi}.}
\tightlist
\item
  Initialize the native planner, an empty target-local verdict cache, and evidence state \(z(R)=\text{unknown}\).
\item
  Select and expand the next molecule using the native backend.
\item
  For every reaction whose structural acquisition trigger fires:

  \begin{enumerate}
  \def\labelenumii{\arabic{enumii}.}
  \tightlist
  \item
    retrieve or query its verifier verdict;
  \item
    if negative, set \(z(R)=\text{negative-pruned}\), mask the action, and propagate the native update;
  \item
    if positive, set \(z(R)=\text{positive-latent}\) unless its positive-release trigger already holds.
  \end{enumerate}
\item
  Under continuous release, recompute each cached positive when its local subtree maturity rises, mark it as released positive evidence, and apply the updated writeback without a new query.
\item
  Under root-delayed release, wait until a root proof closes, extract the route with minimum current effective cost using graph-creation order, and process route reactions in that order. Each positive on this attempted root proof is then released. If a later reaction is negative, that action is pruned and the proof becomes invalid, while earlier released positives remain active and may reprioritize later proofs that reuse them. The negative invalidates the candidate proof, not the positive verdicts already obtained for its other reactions.
\item
  Run a read-only delivery pass that reuses cached observations and queries any remaining route steps. Delivery-only failures do not enter planner state.
\item
  Return the route only if all steps certify; otherwise continue until success or the search budget is exhausted.
\end{enumerate}

Root-delayed release therefore does not require the first closed proof to certify. A failed root proof can still contribute reusable positive evidence from its accepted prefix.

\subsection{Additive-search realization}\label{additive-search-realization}

For retro*, the controller operates on the current minimum-effective-cost proof snapshot. A negative maps its reaction cost to \(+\infty\), invalidates dependent proofs, propagates values to ancestors with the native update routine, removes stale queue entries, and requeues eligible affected molecules. With final released multiplier \(\alpha\) (denoted \(\alpha_{\min}\) in the main text), a released positive maps \(c_0(R)\) to \(\alpha c_0(R)\) and invokes the same propagation. A positive permits continued processing of the current proof snapshot; a negative stops that attempt and forces route reselection. Equal priorities use graph-creation order. The planner is a tree, repeated molecules on a path are prevented, and no hidden verifier score enters the native value model.

\subsection{PUCT realization}\label{puct-realization}

For search state \(s\) and action \(a\), PUCT selects with the native score

\[
Q(s,a)+c_{\mathrm{puct}}P(s,a)\frac{\sqrt{\sum_b n(s,b)}}{1+n(s,a)}.
\]

Here \(Q(s,a)\) is the backed-up mean value, \(n(s,a)\) the visit count, \(P(s,a)\) the policy prior, and \(c_{\mathrm{puct}}\) the exploration constant; \(b\) ranges over sibling actions. Lowercase \(n\) distinguishes PUCT visits from the one-step-call budget \(N\) used elsewhere.

Let \(P_0(s,a)\) be the native policy prior before feedback. Negative evidence masks every existing occurrence of the target-local reaction key, assigns zero prior, and prevents future occurrences from becoming selectable. For an unmasked released positive, VGR uses the unnormalized sibling weight

\[
\omega(s,a)=
\begin{cases}
P_0(s,a)^\alpha, & \alpha>0,\\
1, & \alpha=0,
\end{cases}
\]

and then renormalizes over unmasked siblings; unknown and unreleased positive actions retain native unnormalized weights \(P_0(s,a)\). This piecewise definition matches the maximal-credit endpoint without invoking an undefined \(0^0\). Existing visit counts \(n\), accumulated values, \(Q\), rewards, and the backup rule are unchanged. If every action at a state is masked, that simulation backs up reward zero. The reported configuration uses Syntheseus 0.7.2 \texttt{MolSetMCTS}, \(c_{\mathrm{puct}}=1\), prior temperature 8, constant value 0.5, maximum depth 20, maximum 100,000 simulations, unique-parent nodes, and repeat-molecule prevention. Each target additionally has a 500-call one-step-model budget.

\subsection{Continuous structural maturity}\label{continuous-structural-maturity}

Continuous release uses graph state rather than reaction text or verifier confidence. For a positively verified reaction \(R\) at step \(t\), let \(f_{\mathrm{cl},t}(R)\in[0,1]\) be the fraction of its direct reactant obligations that are topologically closed, \(d_t(R)\in\mathbb N_0\) the number of direct non-stock children whose first expansion consumed a one-step-model call, and \(u_t(R)\in\mathbb N_0\) the number of direct reactant obligations that remain open. We compare

\[
m_{\mathrm{soft},t}(R)=f_{\mathrm{cl},t}(R)
\]

and

\[
\begin{aligned}
m_{\mathrm{inv},t}(R)
&=f_{\mathrm{cl},t}(R)
+\lambda\bigl(1-f_{\mathrm{cl},t}(R)\bigr)\\
&\quad\cdot\frac{d_t(R)}{d_t(R)+u_t(R)+1},
\qquad \lambda=.5.
\end{aligned}
\]

Both definitions are forced to one when the reaction subtree closes. Because pruning and shared subproblems can change the current graph snapshot, each positive stores the greatest maturity it has reached,

\[
\bar m_t(R)=\max_{\tau\le t}m_\tau(R).
\]

For \(\beta>0\), the release fraction is \(r_t(R)=\bar m_t(R)^\beta\), giving

\[
\alpha_{\mathrm{eff},t}(R)
=1-(1-\alpha_{\min})r_t(R).
\]

All continuous experiments fix \(\alpha_{\min}=.25\) and compare \(\beta\in\{1,2\}\). Querying and rescoring are distinct: PI acquisition queries a reaction after the first realized downstream expansion, while later maturity increases reuse the cached verdict and invoke value propagation without another verifier call. The one-way maximum makes released credit monotone; a later negative still prunes the affected action, but maturity fluctuations do not repeatedly add and withdraw credit from a surviving positive action.

\section{Policy Mapping}\label{policy-mapping}

All fixed policies are settings of the same four controls.

\begin{table*}[t]
\centering
\caption{Exact mapping of reported policies to acquisition and activation clocks.}
\small
\resizebox{\textwidth}{!}{%
\begin{tabular}{llllll}
\toprule
Policy & Acquisition & Negative acts & Positive release & Release variable & $\alpha_{\min}$ \\
\midrule
post-hoc & exposed closed route & never in search & never in search & $0$ & -- \\
closure-prune & root proof closure & immediately & neutral & $0$ & 1 \\
closure-guide & root proof closure & immediately & at acquisition & 1 & $[0,1)$ \\
PI-neutral & after first downstream call & immediately & never & $0$ & -- \\
PI-immediate & after first downstream call & immediately & at acquisition & 1 & $[0,1)$ \\
PI-root-delayed & after first downstream call & immediately & root proof closure & $h_t^{\mathrm{root}}(R)$ & $[0,1)$ \\
PI-soft & after first downstream call & immediately & continuous & $\bar m_{\mathrm{soft},t}(R)^\beta$ & .25 \\
PI-investment & after first downstream call & immediately & continuous & $\bar m_{\mathrm{inv},t}(R)^\beta$ & .25 \\
FU-root-delayed & before first downstream call & immediately & root proof closure & $h_t^{\mathrm{root}}(R)$ & .25 \\
\bottomrule
\end{tabular}%
}
\end{table*}

In compact experiment labels, \(\alpha\) denotes the final released multiplier \(\alpha_{\min}\). With that convention, \texttt{prune} is closure-prune with \(\alpha=1\); \texttt{guide} is closure-guide with \(\alpha=0\); \texttt{neutral}, \texttt{immediate}, and \texttt{root-delayed} in the release suite are PI policies with \(\alpha=.25\); \texttt{PI-M} denotes the representative investment-maturity policy with \(\beta=1\); and the shared-ceiling \texttt{closure} arm is closure-guide with \(\alpha=.25\).

The root gate \(h_t^{\mathrm{root}}(R)\) becomes one when the positively verified occurrence is processed on a closed root candidate from the target to stock. Processing stops at the first negative; earlier released positives remain active. It is a global one-way event and is not the same quantity as the time-indexed local reaction-subtree maturity \(m_t(R)\) used by PI-soft and PI-investment.

A genuine downstream expansion is the first expansion of a non-inventory child molecule produced by a reaction that consumes a one-step-model call. Merely generating the reaction is not downstream investment. PI queries just after that call; full-on-use (FU) queries immediately before it and can block the expansion.

\begin{table*}[t]
\centering
\caption{Reproducibility card for the principal datasets and model configurations.}
\label{tab:repro-card}
\small
\begin{tabularx}{\textwidth}{p{.16\textwidth}X p{.23\textwidth}}
\toprule
Component & Identity and transformation & Search use \\
\midrule
Retro190 targets & Public 190-target list distributed with Retro-fallback \citep{tripp2024retrofallback}; canonicalized in the released order & primary mechanism and policy comparison \\
Pistachio-derived targets & 100 DESP Pistachio Hard pairs in source order; canonical targets retained; required starts excluded & target-only transfer with the same stock \\
Inventory & 23,081,583 unique canonical eMolecules entries; zero target-in-stock hits & terminal purchasability test \\
Primary search & template MLP, top-50 applicable sanitized and deduplicated outcomes; learned retro* value; best-first retro*; depth guard 50 & principal environment \\
Transfer search & RetroChimeraDeNovo R-SMILES 2 USPTO-FULL; five root augmentations, beam 50, official parsing and rank aggregation, final top-50; constant-zero value & configuration-level stress test \\
Verifier & ReactionT5v2 forward USPTO-MIT checkpoint; blind reactants; five-beam decoding; canonical non-stereochemical product match & frozen binary feedback and delivery certificate \\
\bottomrule
\end{tabularx}
\end{table*}

\section{Experimental Accounting}\label{experimental-accounting}

\subsection{Shared-ceiling interpretation}\label{shared-ceiling-interpretation}

The \(M_{\mathrm{search}}=100\) and 500 comparisons rerun every policy under a shared ceiling on target-local unique search-time queries. The ceiling equalizes query opportunity, while realized expenditure may differ. Delivery certification remains outside the ceiling, is reported separately, and can reuse search-time cache entries. The trajectory-derived verification-budget curves instead use each run's saved first-certification event and are labeled as observed curves. In the separate \(M_{\mathrm{search}}=500\) rerun, PI-root-delayed uses 43,355 unique search-time queries and closure uses 3,872. The 43,974 PI-root-delayed count reported below belongs to the uncapped release-timing run.

In the uncapped release-timing run, delivery introduced no new unique keys: every route step checked at delivery had already been queried during search. This is why \texttt{search\ unique} and \texttt{cached\ total\ unique} are equal even though delivery performed logical checks.

\begin{table}[h]
\centering
\caption{Query accounting in the uncapped release-timing run.}
\small
\resizebox{\columnwidth}{!}{%
\begin{tabular}{lrrr}
\toprule
Policy & Search unique & Delivery checks & Cached total unique \\
\midrule
PI-neutral & 46,089 & 838 & 46,089 \\
PI-immediate & 49,428 & 901 & 49,428 \\
PI-root-delayed & 43,974 & 969 & 43,974 \\
closure & 3,872 & 889 & 3,872 \\
\bottomrule
\end{tabular}%
}
\end{table}

\subsection{Statistical protocol}\label{statistical-protocol}

Paired target outcomes use exact two-sided McNemar tests. Target-level percentage-point intervals and resource-area intervals use paired bootstrap resampling; FREA uses product-cluster bootstrap resampling. Let \(Y(q,\alpha)\) denote the proportion of targets certified under acquisition clock \(q\) and released positive multiplier \(\alpha\). For the timing--credit experiment, the interaction is \(I=[Y(C,0)-Y(C,1)]-[Y(PI,0)-Y(PI,1)]\), where \(C\) is root-proof closure and \(PI\) is post-investment acquisition. Its intervals use 100,000 target-bootstrap samples and its two-sided interaction tests use target-level sign flips. The fixed-PI release experiment uses 20,000 paired target-bootstrap samples: seed 20261011 for the root-delayed-versus-immediate primary contrast specified before formal analysis, 20261012 for root-delayed versus neutral, 20261013 for neutral versus immediate, and 20261014 for the topology-safety contrast. The two secondary exact tests are corrected as one Holm family: root-delayed versus neutral has raw \(p=.02246\) and adjusted \(p=.04492\); neutral versus immediate has raw and adjusted \(p=.07552\). The full-190 PUCT timing analysis uses 20,000 target-bootstrap samples with seed 20261012. Its full-190 comparison uses one common seed; a separate first-30 pilot uses three common seeds (20261012--20261014) as a repeated-seed robustness check, not as 90 independent target samples. Route-length and proposal-cost directions use paired sign tests. Concurrently served runs are excluded from wall-clock comparisons.

\section{Additional Results}\label{additional-results}

\subsection{Timing--credit interaction}\label{timingcredit-interaction}

We cross two structure-only acquisition clocks with two positive-writeback strengths while holding \(N=500\), \(M_{\mathrm{search}}=100\), negative pruning, the proposal model, and the verifier fixed. Root-proof closure acquires evidence on an exposed complete proof. Post-investment acquisition queries a reaction after the first one-step call spent below it. In both rows, \(\alpha=1\) retains negative pruning but gives no positive credit, whereas \(\alpha=0\) gives maximal positive credit.

Over all 190 targets, closure \(\alpha=1/0\) certifies 113/119 and solves 172/172 topologically; post-investment \(\alpha=1/0\) certifies 116/109 and solves 168/153 topologically. The certification interaction is \(+6.8\) percentage points (95\% CI \([-1.1,+14.7]\)). The topological interaction is \(+7.9\) points (95\% CI \([+4.2,+12.1]\), exact \(p=.0003\)): maximal credit leaves closure topology unchanged but loses 15 post-investment proofs. Thus earlier acquisition and strong guidance are not independently additive improvements. The fixed-PI release experiment in the main paper then isolates positive release and shows that mature positive credit avoids the immediate-credit failure.

\subsection{Continuous-maturity results}\label{continuous-maturity-results}

Continuous local-maturity rules explore an additional release mechanism. Their complete Retro190 results are:

\begin{table*}[t]
\centering
\caption{Continuous structural-maturity controls on Retro190.}
\label{tab:supp-continuous-retro190}
\small
\resizebox{\textwidth}{!}{%
\begin{tabular}{lrrrrr}
\toprule
Policy & Topological & Certified & $\Sigma N$ & Search unique & Node visits \\
\midrule
PI-neutral & 165 & 126 & 44,471 & 46,089 & 299,700 \\
PI-immediate-$\alpha=.25$ & 152 & 116 & 48,758 & 49,428 & 144,272 \\
PI-root-delayed-$\alpha=.25$ & 165 & 135 & 42,579 & 43,974 & 308,057 \\
closure acquisition-$\alpha=.25$ & 172 & 128 & 47,627 & 3,872 & 281,593 \\
soft maturity-$\beta=1$ & 168 & 137 & 42,376 & 43,651 & 262,703 \\
soft maturity-$\beta=2$ & 165 & 135 & 42,310 & 43,553 & 280,331 \\
investment maturity-$\beta=1$ & 169 & 137 & 42,665 & 43,796 & 242,323 \\
investment maturity-$\beta=2$ & 166 & 133 & 42,618 & 43,838 & 279,642 \\
\bottomrule
\end{tabular}%
}
\end{table*}

Soft maturity-\(\beta=1\) and investment maturity-\(\beta=1\) both certify 137 targets. Relative to root-delayed release, each rescues five and loses three (\(+1.05\) points, 95\% CI \([-1.58,4.21]\); \(p=.7266\)), so terminal certification is not separated. Investment-\(\beta=1\) versus closure acquisition is 10 rescued and one lost (\(+4.74\) points, 95\% CI \([1.58,8.42]\); \(p=.01172\)); versus neutral it is 12/1 (\(p=.00342\)), and versus immediate it is 26/5 (\(p=.000192\)).

The second task contains the 100 target molecules from the DESP Pistachio Hard target--required-starting-material pairs. We preserve pair order and retain the original pairs for provenance, but the required starting materials are not supplied to search. Each target is instead planned against the same full eMolecules inventory used for Retro190. We therefore report a Pistachio-derived target-only transfer set rather than the original constrained benchmark.

\begin{table*}[t]
\centering
\caption{Continuous structural-maturity controls on the 100-target Pistachio-derived target-only set.}
\label{tab:supp-continuous-pistachio}
\small
\resizebox{\textwidth}{!}{%
\begin{tabular}{lrrrrr}
\toprule
Policy & Topological & Certified & $\Sigma N$ & Search unique & Node visits \\
\midrule
post-hoc & 82 & 61 & 22,804 & 0 & 129,969 \\
neutral & 77 & 64 & 18,652 & 18,837 & 120,860 \\
immediate-$\alpha=.25$ & 79 & 66 & 17,859 & 17,941 & 59,503 \\
root-delayed-$\alpha=.25$ & 77 & 66 & 18,053 & 18,243 & 131,613 \\
soft maturity-$\beta=1$ & 80 & 67 & 17,331 & 17,496 & 113,247 \\
soft maturity-$\beta=2$ & 79 & 69 & 17,513 & 17,758 & 122,458 \\
investment maturity-$\beta=1$ & 80 & 69 & 17,136 & 17,381 & 102,735 \\
investment maturity-$\beta=2$ & 79 & 68 & 17,436 & 17,654 & 115,858 \\
\bottomrule
\end{tabular}%
}
\end{table*}

Investment-\(\beta=1\) rescues three targets and loses none relative to root-delayed release (\(+3\) points, 95\% CI \([0,7]\); \(p=.25\)), while using 917 fewer one-step calls, 862 fewer unique queries, and 28,878 fewer node visits. Relative to post-hoc it rescues eight and loses none (\(p=.0078\)). On Retro190, every continuous instance avoids the large immediate-credit drop. Across both tasks, these rules provide local, interpretable alternatives to the global root-release gate.

\subsection{Shared ceilings and the acquisition frontier}\label{shared-ceilings-and-the-acquisition-frontier}

Shared-ceiling reruns distinguish equal maximum query opportunity from trajectory-derived resource curves:

\begin{table}[h]
\centering
\caption{Certified targets under shared target-local unique search-query ceilings.}
\label{tab:supp-shared-caps}
\small
\resizebox{\columnwidth}{!}{%
\begin{tabular}{lrrrr}
\toprule
$M_{\mathrm{search}}$ & PI-neutral & PI-immediate & PI-root-delayed & Closure \\
\midrule
100 & 118 & 112 & 113 & 126 \\
500 & 126 & 115 & 135 & 128 \\
uncapped & 126 & 116 & 135 & 128 \\
\bottomrule
\end{tabular}%
}
\end{table}

At \(M_{\mathrm{search}}=100\), closure acquisition exceeds PI-root-delayed 126--113 (18 rescues, five losses; \(p=.0106\)). At 500, PI-root-delayed exceeds closure acquisition 135--128 (\(p=.0391\)), immediate 135--115 (\(p=.000535\)), and neutral 135--126 (\(p=.0225\)). A shared cap equalizes opportunity, not expenditure: PI-root-delayed makes 43,355 unique search-time queries in the \(M=500\) rerun, whereas closure acquisition makes 3,872.

The wider acquisition sweep uses logical-event limits: post-hoc has no search feedback, closure uses \(M_{\mathrm{event}}=100\), and early-prune uses \(M_{\mathrm{event}}=500\). PI-root-delayed and FU-root-delayed have no active event ceiling. If \(S(n)\) is the certified-target fraction by one-step-call budget \(n\), we report \(\mathrm{AUSC}@500=500^{-1}\sum_{n=1}^{500}[S(n-1)+S(n)]/2\), the normalized trapezoidal area through all integer budgets from 0 to 500.

\begin{table*}[t]
\centering
\caption{Observed search--verification frontier under the reported acquisition sweep.}
\label{tab:supp-frontier}
\small
\resizebox{\textwidth}{!}{%
\begin{tabular}{lrrrrrrr}
\toprule
Policy & Certified & Topological & $\Sigma N$ & AUSC@500 & Feedback events & Cached unique & Positive maturation \\
\midrule
post-hoc & 97 & 172 & 57,365 & .397 & 0 & 3,502 & -- \\
closure-$\alpha=.25$ & 121 & 172 & 48,970 & .473 & 4,715 & 3,790 & -- \\
early-prune & 121 & 165 & 46,416 & .497 & 46,067 & 47,803 & -- \\
PI-root-delayed-$\alpha=.25$ & 135 & 165 & 42,579 & .537 & 56,169 & 43,974 & 28.0\% \\
FU-root-delayed-$\alpha=.25$ & 144 & 148 & 34,580 & .621 & 416,570 & 99,399 & 1.19\% \\
\bottomrule
\end{tabular}%
}
\end{table*}

Here early-prune uses PI acquisition, immediate negative pruning, and neutral positives. Positive maturation is the number of reaction occurrences that receive released positive credit divided by the number receiving positive evidence during search; distinct occurrences sharing a cached reaction key count separately. Aggregated over targets, these ratios are \(8{,}496/30{,}328\) for PI and \(2{,}812/235{,}994\) for FU.

PI versus closure rescues 15 targets and loses one (\(p<.001\)); FU versus closure rescues 27 and loses four (\(p<.001\)); FU versus PI rescues 14 and loses five (\(p=.064\)). FU uses 7.42 times the feedback events and 2.26 times the unique queries of PI, while only 1.19\% of its positive observations mature into released credit. These policies occupy different points on a two-resource frontier: earlier acquisition exchanges more verifier work for fewer proposal calls.

\subsection{Transfer statistics}\label{transfer-statistics}

With RetroChimera plus retro*, no feedback, negative-only feedback, and positive-plus-negative feedback certify 162, 174, and 180 targets. Positive-plus-negative feedback rescues seven and loses one target relative to negative-only feedback (exact McNemar \(p=.0703\)). With the original proposal model and PUCT, the corresponding counts are 91, 113, and 121; positive-plus-negative feedback rescues twelve and loses four targets (\(p=.0768\)). These endpoint tests are directionally consistent but individually inconclusive; they test interface transfer rather than establish a new significant coverage effect.

The full-190 PUCT timing experiment uses a common seed and fixes PI acquisition and immediate negative masking. Here \(\mathrm{AUSC\mbox{-}N}@500=501^{-1}\sum_{n=0}^{500}S_N(n)\), where \(S_N(n)\) is the certified-target fraction by one-step call \(n\); \(\mathrm{AUSC\mbox{-}sim}@100\mathrm{k}\) is defined analogously over simulation budgets 0 through 100,000. Its results are:

\begin{table}[h]
\centering
\caption{Full-190 PUCT coverage and one-step-call results.}
\label{tab:puct-timing}
\small
\resizebox{\columnwidth}{!}{%
\begin{tabular}{lrrrr}
\toprule
PUCT policy & Certified & Topological & $\Sigma N$ & AUSC-N@500 \\
\midrule
PI-neutral & 132 & 151 & 39,318 & .569 \\
PI-immediate-$\alpha=.25$ & 132 & 148 & 38,778 & .573 \\
PI-root-delayed-$\alpha=.25$ & 136 & 151 & 37,465 & .590 \\
PI-immediate-$\alpha=0$ & 125 & 143 & 39,492 & .564 \\
PI-root-delayed-$\alpha=0$ & 137 & 151 & 37,009 & .593 \\
\bottomrule
\end{tabular}%
}
\end{table}

\begin{table}[h]
\centering
\caption{Full-190 PUCT simulation and graph results.}
\small
\resizebox{\columnwidth}{!}{%
\begin{tabular}{lrrr}
\toprule
PUCT policy & $\Sigma$ simulations & AUSC-sim@100k & $\Sigma$ nodes \\
\midrule
PI-neutral & 926,369 & .687 & 9,595,796 \\
PI-immediate-$\alpha=.25$ & 1,126,527 & .685 & 18,520,797 \\
PI-root-delayed-$\alpha=.25$ & 866,633 & .708 & 8,725,386 \\
PI-immediate-$\alpha=0$ & 1,315,355 & .652 & 22,687,335 \\
PI-root-delayed-$\alpha=0$ & 855,568 & .714 & 8,611,078 \\
\bottomrule
\end{tabular}%
}
\end{table}

Root-delayed-\(\alpha=0\) versus immediate-\(\alpha=0\) gives 13 rescues and one loss (\(p=.001831\)); root-delayed versus neutral gives five rescues and no losses (\(p=.0625\)). The latter endpoint contrast is inconclusive, while its paired AUSC differences on the one-step-call and simulation axes have target-bootstrap 95\% intervals \([+.0091,+.0406]\) and \([+.0061,+.0518]\). All curves use saved first-certification points; concurrent wall-clock is excluded.

Before the full-190 run, the first-30 pilot repeated the same targets under three common seeds. Across the resulting 90 paired target--seed runs, PI-neutral, PI-immediate-\(\alpha=.25\), PI-root-delayed-\(\alpha=.25\), PI-immediate-\(\alpha=0\), and PI-root-delayed-\(\alpha=0\) produced 78, 75, 78, 72, and 78 verifier-certified outcomes, respectively. These repetitions were a robustness check; inferential tests in the paper keep the target, rather than each repeated run, as the sampling unit.

\subsection{External evidence for the route verifier}\label{external-evidence-for-the-route-verifier}

On 1,000 expert-labeled URSA reactions, the primary verifier obtains 77.1\% balanced accuracy (95\% CI \([74.5,79.6]\)) and Matthews correlation coefficient (MCC) 0.547, with 418 true positives, 353 true negatives, 147 false positives, and 82 false negatives. Independently predicted reaction conditions raise balanced accuracy to 79.6\%. Graph2SMILES obtains 74.9\% balanced accuracy on the same set.

On the held-out FREA collection, the verifier is evaluable on 740 of 751 reactions and obtains 71.4\% balanced accuracy, with a product-cluster bootstrap 95\% CI of \([67.8,74.9]\).

The fixed-route PaRoutes \citep{genheden2022paroutes} audit uses the 251 targets for which guide succeeded under an online Graph2SMILES verifier. The corresponding guide and first-topological routes were frozen before cross-model rescoring. Within that guide-success cohort, a condition-aware ReactionT5 model absent from online selection accepts 181/251 guide routes and 117/251 first-topological routes. The paired difference is \(+25.5\) percentage points (95\% CI \([18.7,31.5]\)), with 73 guide-only and 9 base-only acceptances (exact McNemar \(p<10^{-12}\)). Mean route length changes from 3.77 to 5.08 reactions. These estimates describe the selected paired cohort.

\section{Reproducibility and Limitations}\label{reproducibility-and-limitations}

Table \ref{tab:repro-card} summarizes the main experimental identities and transformations. Exact file identities and complete manifests are retained with the experiment archive but are not used as scientific claims. The verifier environment uses PyTorch 2.8.0+cu128, Transformers 4.40.2, CUDA 12.8, deterministic execution, TF32 disabled, and batch size one. The experiment archive records target-level outcomes, first-certification events, route records, separate search and delivery counters, random seeds, software versions, and file manifests. The PUCT configuration and statistical seeds are given above.

Two experimental gaps remain. The current shared-ceiling reruns cap search-time queries but do not match total realized search-plus-delivery verifier cost; a total-query-budget intervention would strengthen the resource analysis. In addition, the verifier has nonzero expert-label error and no wet-lab synthesis is reported. Holding out external evaluation labels prevents direct tuning on them, while possible overlap in the underlying training corpora remains unmeasured.
\end{document}